\documentclass{jmlr}
\jmlrvolume{}
\jmlryear{}
\jmlrproceedings{}{}
\editor{}
\jmlrworkshop{}
\jmlrpages{}

\usepackage{arxiv27-causality-rscience-frame}
\usepackage{arxiv27-causality-rscience-values}

\graphicspath{{figures}}

\begin{document}
\title{Mining Natural Language Internet Data for Causality}

\author{%
\Name{Tim Hagen} \Email{tim.hagen@uni-kassel.de}\\
\addr University of Kassel and hessian.AI%
\AND
\Name{Liesbeth Allein}
\Email{Liesbeth.Allein@ugent.be}
\\
\addr Ghent University%
\AND
\Name{Martin Potthast} \Email{martin.potthast@uni-kassel.de}\\
\addr University of Kassel, hessian.AI, and ScaDS.AI%
}

\maketitle

\begin{abstract}
Observational data alone often cannot distinguish between competing causal structures. Background knowledge can help constrain causal discovery and inform the formulation and evaluation of causal hypotheses. Much of this knowledge, however, is expressed in natural language, where causal relationships are reported as claims rather than established facts. Such claims may be incomplete, conflicting, and expressed at different levels of specificity. We introduce a large-scale hierarchical graph of causal claims that makes this information accessible without imposing a single causal model. We propose a modular pipeline for constructing the graph from \numtitles{} submission titles from Reddit's r/science community. The pipeline extracts causal claims and counterclaims, canonicalizes event descriptions into self-contained representations, and organizes related events hierarchically at different levels of abstraction. Each claim retains its textual provenance, allowing assertions to be traced to their sources. The resulting graph contains \numaftercanonicalization{} causal and countercausal relationships across scientific domains.
We evaluate the quality of extraction, canonicalization, and hierarchical organization.
By making reported causal relationships retrievable and comparable while preserving disagreement and provenance, the graph provides a basis for acquiring candidate background knowledge, investigating competing causal hypotheses, and developing methods for reasoning over incomplete or conflicting causal information.%
\footnote{Code: \url{https://github.com/webis-de/arxiv-causality-rscience}; data will be made available later.}
\end{abstract}

\section{Introduction}
Formal causal learning and reasoning seek to characterize causal relationships between variables and events. Statistical approaches typically operate on structured observations under assumptions about the underlying causal system~\citep{scholkopf:2021}. Yet, observational data may leave several causal structures indistinguishable within a Markov equivalence class. While background knowledge can distinguish these further~\citep{perkovic:2017,kiciman:2024,bang:2025}, much potentially relevant knowledge is expressed in natural language: scientific publications, reports, and public discourse contain findings, hypotheses, and disagreements about causes and effects. In industrial root-cause analysis, for example, causal discovery from sensor data~\citep{mehling:2026} and extraction of reported causes from maintenance records~\citep{hershowitz:2024} offer complementary approaches to identifying potential causes of equipment failures.

However, natural language claims about causality (\emph{causal claims}) cannot be directly interpreted as formal causal models. A maintenance report stating that motor overheating causes pump failure is not an identified causal effect. Other reports may dispute causality through \emph{countercausal claims} (e.g., motor overheating does not cause pump failure). Deciding which claim to accept requires additional evidence or assumptions.
Claimed causal relationships also may not correspond to direct causal dependencies and could depend on context omitted by the writer for brevity~(maxim of quantity~\citep{grice:1975}). Thus, treating such claims as edges of a single causal DAG adds assumptions not established by the sources.

Graphs of causal claims must therefore be distinguished from causal models, which specify assumptions about causal relationships in an underlying system. Structural causal models (SCMs) define causal dependencies through structural equations and provide formal semantics for interventions and counterfactuals. Qualitative and nonmonotonic causal theories offer alternative formalizations, with established connections to SCMs~\citep{bochman:2023}. The distinction is therefore not between qualitative and quantitative causality, but between representing causal assertions and modeling a consistent causal theory. Causal-claim graphs preserve the former without committing to a particular causal theory and can inform the construction of one~(Table~\ref{tab:claimgraph-vs-scm}).
\begin{table*}
\centering
\small
\tabcolsep=3.1pt
\renewcommand{\arraystretch}{.9}%
\belowrulesep=\dimexpr\arraystretch \belowrulesep\relax%
\aboverulesep=\dimexpr\arraystretch \aboverulesep\relax%
\belowbottomsep=\belowrulesep%
\caption{Comparison of hierarchical causal-claim graphs with provenance and SCMs.}
\begin{tabular}{@{}l*{2}{p{6.338cm}}@{}}
\toprule
     & \centerline{\textbf{Causal-claim Graph}} & \centerline{\textbf{Causal DAG / SCM}} \\
\midrule
\textbf{Represents}
    & Source-attributed causal claims.
    & Assumed causal relationships within a model. \\\noalign{\smallskip}
\textbf{Causal notion}
    & Does not prescribe a formal causal interpretation of the reported claims.
    & SCMs define interventions and counterfactuals through structural equations. \\\noalign{\smallskip}
\textbf{Edge meaning}
    & Source $s$ claims that $A$ [not] causes $B$.
    & $A$ is modeled to directly cause $B$.\\\noalign{\smallskip}
\textbf{Structure}
    & May contain conflicting claims and cycles; organizes event descriptions hierarchically.
    & A causal DAG is acyclic; an SCM specifies structural dependencies between variables. \\\noalign{\smallskip}
\textbf{Illustration}
    & \raisebox{-\height}{\centerline{\includegraphics{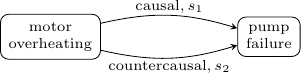}}}
    & \raisebox{-\height}{\centerline{\includegraphics{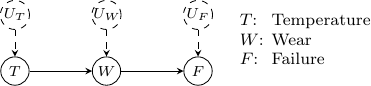}}}\\
\bottomrule
\end{tabular}
\label{tab:claimgraph-vs-scm}
\end{table*}

Such graph representations are relevant to several problems in causal learning and reasoning. Background knowledge can constrain discovery~\citep{bang:2025}, while disagreement about causal relationships and uncertainty about the structure motivate reasoning about alternative causal models~\citep{padh:2025,baldo:2026}. Causal-claim graphs make candidate causal relationships retrievable and comparable while preserving opposing assertions and their provenance and thus could support the formulation and evaluation of causal hypotheses, can support evidence-grounded causal question answering, and provide input for qualitative reasoning over incomplete or conflicting causal knowledge~\citep{cui:2024a,hagen:2025a,allein2026implicit}.

Constructing causal-claim graphs from text presents several challenges. Causality can be expressed through diverse lexical, syntactic, and even purely semantic constructions~\citep{altenberg:1984,girju:2002,prasad:2008,allein:2026}. Extracting countercausal claims is particularly important to reduce bias towards causal claims~\citep{hagen:2025a}. Furthermore, extracted event expressions may omit contextual details needed for independent interpretation. Across sources, the same event may be described differently, while related events may differ in specificity. A claim about a particular event may not generalize to a broader category or can be overridden by more specific counterclaims. Integrating such claims therefore requires preserving their meaning and provenance while organizing related event descriptions without conflating them.

\paragraph{Contributions} We create \emph{a large-scale causal-claim graph} that jointly
(1)~represents both causal claims and counterclaims,
(2)~canonicalizes event descriptions into self-contained descriptions, and
(3)~organizes events and concepts hierarchically at different levels of abstraction, while
(4)~each causal claim retains its textual provenance.
We construct the graph from \numtitles{} submission titles on Reddit's r/science community, which provides records of findings across various scientific domains. The final graph contains \numaftercanonicalization{} counter-/causal relationships and provides a resource for studying causal knowledge retrieval, causal question answering, and reasoning over qualitative causal claims. Note that the pipeline we develop is very modular and is not specific to r/science, which makes it applicable in principle to other textual sources beyond scientific discourse.
\section{Related Work}

\paragraph{Causality Extraction from Natural Language}
Causality extraction identifies causal information expressed in natural language text through three subtasks~\citep{tan:2023a}: (1) detecting whether a text contains causal information, (2) extracting the participating cause and effect spans, and (3) identifying the causal relationship between them. Earlier neural approaches model these tasks using text classification and sequence tagging, sometimes incorporating external knowledge~\citep{li:2019,li:2021a,khetan:2021}. Large language models (LLMs) have been increasingly used for causality extraction~\citep{gao-etal-2023-chatgpt,sun-etal-2024-event,liu-etal-2025-large-language}. Nonetheless, fine-tuned transformer models remain competitive, particularly for implicit and complex causal expressions~\citep{takayanagi:2024,Norouzi_Kleinberg_Vermunt_van_Lissa_2025}. We test and compare LLM-based and fine-tuned methods throughout our pipeline.

Prior work gave little attention to countercausality and event canonicalization. This is reflected in the resources available for finetuning. Datasets predominantly capture claims that assert causal relationships only \citep{mihaila:2013,mirza:2014a,dunietz:2017,tan:2022a,tan:2023a,romanou-etal-2023-crab,kunle-john-etal-2026-pubmedcausal}. Negated relationships are often removed explicitly. Yet, explicitly representing counterclaims is important for reasoning over incomplete and potentially conflicting causal knowledge extracted from text. \citet{hagen:2025a} created a dataset focused on countercausality and showed that models trained without countercausal examples tend to misclassify them as causal. Regarding event canonicalization, the large majority of resources did not perform any normalization. Those that did, did this manually and enforced a strict noun phrase structure on the canonicalized event \citep{allein:2026,allein2026implicit}. We finetune and evaluate the extraction methods in our pipeline on data capturing causal and countercausal claims. We further add a canonicalization step with noun phrase completion and coreference resolution with deterministic clause realization, with a fallback to constrained LLM generation.

\paragraph{Causal Claim Graphs} Previous work extracted causal relationships from text and organized them into structured resources~\citep{girju:2002,mihaila:2013,mirza:2014a} or even from LLMs' parametric knowledge~\citep{hu:2025}. We follow the former approach, and Table~\ref{tbl:causal-graphs} compares our causal claim graph to other text-sourced causal graphs~\citep{luo:2016,li:2020,heindorf:2020,LYU2023104298,xu2025fincakg,garg:2025} and more general knowledge bases~\citep{rashkin:2018,mostafazadeh:2020}. Larger causal knowledge graphs like CausalNet~\citep{luo:2016}, Cause Effect Graph~\citep{li:2020}, CauseNet~\citep{heindorf:2020}, and Clinical CKG~\citep{LYU2023104298} extracted causal relationships through rule-based patterns. They use predefined lists of positive causal cues (e.g., \textit{cause}, \textit{due to}) and causal counterclaims were typically filtered. Using a (large) language model-based pipeline for causality extraction, FinCaKG~\citep{xu2025fincakg} and Economics Claim Graph~\citep{garg:2025} account for implicit causality. However, these graphs also do not represent counterclaims. Clinical CKG~\citep{LYU2023104298} normalized entities to UMLS concepts and related clinical terminologies. By contrast, we canonicalize event spans as self-contained natural-language descriptions that retain claim-specific context. Instead of imposing a predefined ontology-based hierarchy, we infer subsumption relations between extracted events and use existing ontologies to supply further general concepts.

\begin{table}%
\centering\small%
\tabcolsep=3.9pt%
\renewcommand{\arraystretch}{.9}%
\belowrulesep=\dimexpr\arraystretch \belowrulesep\relax%
\aboverulesep=\dimexpr\arraystretch \aboverulesep\relax%
\belowbottomsep=\belowrulesep%
\newcommand{\algn}[2]{\hspace{\widthof{0} * #1 + \widthof{\,} * \numexpr(#1-1)/3\relax}\makebox[0pt][r]{#2}}%
\caption{Comparison of graphs or knowledge graphs with causal information.}
\begin{tabular}{@{}p{7.0cm}l@{}cc@{\hspace{.75\tabcolsep}}c@{\hspace{.75\tabcolsep}}c@{\hspace{.75\tabcolsep}}c@{\hspace{.75\tabcolsep}}c@{}}
\toprule
\textbf{Graph}             & \textbf{Domain} & \textbf{\# Causal Rels.} & \textbf{P} & \textbf{I} & \textbf{CC} & \textbf{CA} & \textbf{H} \\
\midrule
\multicolumn{8}{@{}l@{}}{\footnotesize\itshape Knowledge bases with causal knowledge (among others)}      \\
Event2Mind\ \citep{rashkin:2018}                 & Social; mental states &     \algn{6}{25\phantom{.0}}\,K\makebox[0pt][l]{*} & $\checkmark$  & $\checkmark$  & --   & -- &  -- \\
GLUCOSE\ \citep{mostafazadeh:2020}                    & Narrative &     \algn{6}{335\phantom{.0}}\,K\makebox[0pt][l]{*} & $\checkmark$  & $\checkmark$ & --  & -- & -- \\
\midrule
\multicolumn{8}{@{}l@{}}{\footnotesize\itshape Causal knowledge graphs}   \\
CausalNet\ \citep{luo:2016}                  & General &     \algn{6}{62.7}\,M & --  & --  &  -- & -- & -- \\
Cause Effect Graph\ \citep{li:2020}          & General &     \algn{6}{89.1}\,M & --  & --  &  --  & -- &  -- \\
CauseNet\ \citep{heindorf:2020}              & General &     \algn{6}{11.6}\,M & $\checkmark$ & -- &  -- & -- & -- \\
Clinical CKG\ \citep{LYU2023104298}          & Medicine & \algn{6}{1.7}\,M & \checkmark & -- & -- & \checkmark & \checkmark \\ 
FinCaKG\ \citep{xu2025fincakg}               & Finance & \algn{6}{11.6}\,K & -- & $\checkmark$ & -- & -- & -- \\ 
Economics Claim Graph~\citep{garg:2025}      & Economics & \multicolumn{1}{c}{\textit{Not reported}} & $\checkmark$ & $\checkmark$ & -- & -- & -- \\
\midrule
\textbf{Ours}                                & Science; multi-domain & \algn{6}{198.6}\,K & $\checkmark$ & $\checkmark$ &  $\checkmark$ & $\checkmark$ & $\checkmark$ \\
\bottomrule
\multicolumn{8}{@{}p{\linewidth}@{}}{
\textbf{P}:~textual or source provenance;
\textbf{I}:~implicit causality;
\textbf{CC}:~causal counterclaims;
\textbf{CA}:~canonized events;
\textbf{H}:~an explicit hierarchy connecting event descriptions.
$^*$~Relations are not exclusively causal.}
\end{tabular}
\label{tbl:causal-graphs}
\end{table}

\section{Method}
This section describes the data source (Section~\ref{sec:datasource}) and causality extraction pipeline~(Section~\ref{sec:pipeline}). Note that the pipeline is not specific to this paper's data source and can identically be applied to other data sources and domains.

\subsection{Data}
\label{sec:datasource}
We retrieve submission titles from Reddit's r/science community, obtained from the Pushshift Reddit dataset~\citep{baumgartner:2020}. This subreddit is dedicated to sharing and discussing scientific research across disciplines. Its submission rules\footnote{\url{https://www.reddit.com/r/science/wiki/rules}} require links to peer-reviewed research or their professional summaries. They also discourage editorialized, sensationalized, and biased submission titles. These titles typically summarize the main findings of the linked sources in concise natural language. When findings contain causal descriptions, the titles provide short accounts of causal relationships, although qualifications and contextual information from the underlying research may be omitted. The corpus comprises \numprefilter{} submissions published between 2006 and 2022.

\paragraph{Preprocessing}
We remove duplicate titles and deleted or removed submissions. Duplicates are identified after Unicode normalization, case folding, and whitespace normalization. We retain Ask Me Anything (AMA) submissions as some contain causal claims in their titles. In total, we remove \numduplicatesremoved{} duplicate titles and \numplaceholdersremoved{} deleted posts and keep \numtitles{} titles for further processing.

\paragraph{Corpus Characteristics}
Submission titles provide concise textual accounts of reported scientific findings; mean length is \titlewordsmean{} words (median: \titlewordsmedian{}; 5th--95th percentile: \titlewordsplow{}--\titlewordsphigh{}). We characterize the disciplinary distribution by the topic labels (\emph{flairs}) assigned to submissions by their authors. We normalize these labels and manually map them to the scientific fields of the OpenAlex taxonomy \citep{priem:2022}. We also explored automated topic classification, but preliminary experiments indicated that these approaches introduced additional noise. We therefore rely on the author-assigned flairs.
Since authors sparsely assigned flairs in earlier years, we restrict the topic analysis to 2019--2022, when flair coverage increased from 84\% to nearly 100\% (93.3\% overall). This period contains approximately 14\% of the retained corpus. The restriction applies only to the descriptive topic analysis; all available years are included in the causal-claim extraction pipeline.

\begin{figure}
\centering
\includegraphics{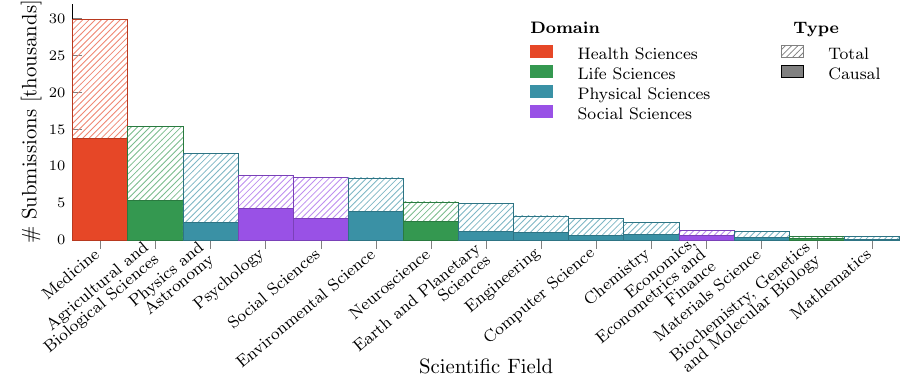}
\caption{The distribution of topics in the r/science subreddit titles from 2019 until 2022. Dashed bars mark the overall distribution whereas the solid bars mark the topic distribution of titles with extracted causal information.}
\label{fig:topic-distribution}
\end{figure}

Figure~\ref{fig:topic-distribution} shows the distribution across the OpenAlex fields%
\footnote{\url{https://openalex.org/fields}}. It groups the fields into four broader scientific domains and shows the number and proportion of titles from which our pipeline extracted at least one (counter-)causal relationship. \emph{Medicine} is the most prominent field in both submissions and extracted claims, with nearly half of its titles yielding causal information. \emph{Agricultural and Biological Sciences} follows. \emph{Physics and Astronomy} ranks third by submission count, but ranks only eighth by the number of extracted claims. Conversely, \emph{Psychology} and \emph{Environmental Science} have fewer submissions but contribute about twice as many extracted claims. These results demonstrate that the prevalence of (extractable) causal information differs substantially across scientific fields.

\subsection{Causality Extraction Pipeline}
\label{sec:pipeline}
\begin{figure}
\centering
\includegraphics{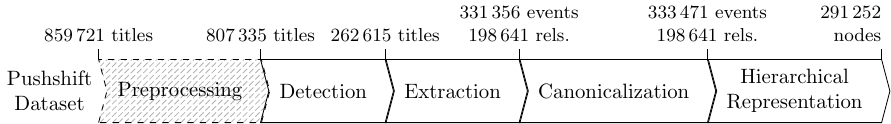}
\caption{The causality extraction pipeline we use to create the causal graph.}
\label{fig:pipeline}
\end{figure}

We design a four-step pipeline (Figure~\ref{fig:pipeline}):
(1)~\emph{detection} classifies whether a given piece of text contains causal information;
(2)~\emph{extraction} extracts all causal claims and counterclaims contained in that piece of text and identifies the event spans;
(3)~\emph{canonicalization} normalizes those spans to make event descriptions more self-contained; and
(4)~\emph{hierarchical representation} organizes event descriptions hierarchically. The pipeline is modular, i.e., the methods supporting each step are interchangeable.
\label{sec:datasets}
We use CCNC~\citep{hagen:2025a}, CNCv2~\citep{tan:2023}, and BECauSE 2.0~\citep{dunietz:2017} for fine-tuning and evaluating detection and extraction methods. CCNC is particularly important since it is the only dataset with annotated countercausality. All three datasets follow the same annotation scheme~\citep{grivaz:2010}, which aligns well with the kind of causal claims that should be extracted here and how spans should be selected. In preliminary experiments, we tested additional datasets with different annotation approaches. However, combining these datasets reduced the model's effectiveness. For CCNC and CNCv2, the official test sets are not publicly available. We therefore use their official validation sets as held-out test sets and split the original training sets into new training and validation partitions. For BECauSE 2.0, we adopt the train/test split by UniCausal~\citep{tan:2023a} and reserve random 10\% of its training data for validation. See Table~\ref{tab:dataset-statistics} in the Appendix for dataset statistics.

\paragraph{Step 1 (Detection)}
We classify each submission title as containing causal information (\emph{causal}) or not (\emph{uncausal}). We fine-tune RoBERTa-base~\citep{liu:2019} as a binary classifier on the datasets described above.
Recall is prioritized here since causal information in titles classified as uncausal cannot be recovered by subsequent pipeline steps. We therefore use a confidence threshold of 0.5 for the causal class. On the CCNC validation set, e.g., increasing this threshold to 0.7 reduced recall from 0.892 to 0.881 (see Appendix~\ref{sec:detection-threshold} for more detailed results). We retain the lower threshold to avoid discarding potentially relevant claims, while allowing the next step to filter false positives.

\paragraph{Step 2 (Extraction)}
For each title classified as causal, we extract claimed causal and countercausal relationships as $\langle\texttt{cause span},\texttt{effect span},\texttt{relation}\rangle$-tuples. Step 2 consists of two subtasks: \emph{candidate extraction}, which identifies textual event spans that may participate in causal relationships, and \emph{causality identification}, which classifies the relationship between ordered pairs of candidate spans. \emph{Candidate extraction} relies on a fine-tuned RoBERTa-base~\citep{liu:2019} with a biaffine span-prediction head~\citep{yu:2020}; the encoder produces contextualized token representations from which candidate start and end positions are scored directly. In preliminary experiments, this approach marked span boundaries more precisely and is less prone to fragmenting spans and merging adjacent event descriptions than an encoder transformer fine-tuned for token classification using BIO labels~\citep{ramshaw:1995}.
Note that precise event boundaries can be difficult to establish, even for human annotators~\citep{tan:2023,colruyt:2019}. Our pipeline therefore does not require exact span matches, but aims to identify the relevant event expressions. These are later expanded using their source context during canonicalization.
We select a span confidence threshold of 0.5 based on precision--recall analysis on the combined validation sets (see Appendix~\ref{sec:candidate-extraction-threshold} for more information). For example, increasing the threshold to 0.8 improves character-overlap precision from 0.728 to 0.756 but reduces recall from 0.829 to 0.801, while character-overlap F\textsubscript{1} improves only marginally (0.775 to 0.778). We therefore favor the higher-recall operating point to retain potentially relevant event spans for subsequent identification and canonicalization.
\emph{Causality identification} relies on a fine-tuned Ettin decoder 150\,M model~\citep{weller:2026}. Given the original title with two candidate spans explicitly marked, the model classifies their directed relationship as \emph{causal}, \emph{countercausal}, or \emph{uncausal}. We retain pairs classified as causal or countercausal and discard uncausal pairs. Multiple relationships may be extracted from the same title. Coordinated expressions such as ``A causes B and C'' should yield a single relationship. Since the training data for causality extraction already marks conjunctions as separate relationships, it is not necessary to perform additional conjunction splitting.

\paragraph{Step 3 (Canonicalization)}
We normalize extracted event spans into independently interpretable descriptions while preserving modifiers that determine the causal claims' scope. A manual linguistic analysis of 324 cause and effect spans across 161 domain-stratified titles identified the importance of agents and affected entities~\citep{fillmore:2003} and spatiotemporal context. For example:
\begin{adjustwidth}{2.8em}{0pt}
    \textbf{Agent}: In \textit{``\underline{White America’s racial resentment}\textsubscript{\texttt{cause}} is the real impetus for welfare cuts, study says''}, excluding the agent (\textit{``White America''}) from the formulation of the cause underspecifies the cause and removes meaningful responsibility attribution;\\
    \textbf{Affected entity}: In \textit{``\underline{Altitude adaptation in Tibetans}\textsubscript{\texttt{effect}} caused by introgression of De\-ni\-so\-van-like DNA''}, the affected entity (\textit{``Tibetans''}) signals the population for which the effect (\textit{``altitude adaptation''}) is asserted. It may also imply the existence of confounding variables. It is therefore crucial to include it in the formulation of the effect; \\
    \textbf{Spatiotemporal contextualization}: In \textit{``Study finds \underline{peanut consumption in infancy}\textsubscript{\texttt{cause}} may prevent peanut allergy''}, excluding time indication (\textit{``in infancy''}) from the cause formulation would create a negative causal relationship (i.e., prevention)~\citep{goldvarg2001naive} that is nonsensical and goes right against the observed causal relation.
\end{adjustwidth}
We therefore canonicalize each event independently, using its source title as context without incorporating information from the paired event. Our modular pipeline supports syntactic expansion, context-aware canonicalization, and LLM-based rewriting. Syntactic expansion completes noun phrases using dependency relations, including possessive modifiers, prepositional phrases, and relative clauses. Context-aware canonicalization additionally resolves coreferences and converts event phrases into self-contained clauses using deterministic rules. A genericity check preserves general concepts as phrases where sentence realization would introduce unnecessary information. The generated descriptions are checked for semantic preservation using content-word overlap and for leakage from the paired event; otherwise, the original or expanded phrase is retained. We use this context-aware approach for the released graph. LLM-based rewriting with constrained generation is evaluated separately but not used in the final representation (Appendix~\ref{sec:llm-canonization}).

\paragraph{Step 4 (Hierarchical Representation)} Extracted event descriptions vary substantially in specificity. For example, \textit{``injury''}, \textit{``severe injury''}, and \textit{``spinal cord injury''} should remain distinct, while claims about injuries in general may also be relevant when investigating spinal cord injuries. Organizing these descriptions hierarchically makes related claims accessible at different levels of abstraction without conflating their meanings. Specificity also matters for potentially conflicting causal claims. Consider \textit{``fire causes smoke''} and \textit{``methane fire does not cause smoke.''} If the former expresses a general causal default, the latter may represent a more specific exception. Human causal reasoning accounts for conditions under which expected effects fail to occur~\citep{cummins:1995}, while default reasoning allows general expectations to be revised by more specific information~\citep{pearl:1988,poole:1991}. Since natural-language causal claims often leave such qualifications implicit~\citep{grice:1975,hagen:2025a}, we organize event descriptions hierarchically without imposing additional causal assumptions.

We represent hierarchical relationships through semantic subsumption: We write $A\sqsubseteq B$ to denote that every event described by $A$ is also an instance of the more general description $B$\footnote{We use $A\sqsubset B$ for $A\sqsubseteq B\land \lnot (B\sqsubseteq A)$, and we use $\sqsupseteq$ and $\sqsupset$ as the converse relations to $\sqsubseteq$ and $\sqsubset$, respectively.}.
We infer these relationships using natural language inference~(NLI), which assesses whether one text entails another~\citep{dagan:2005}. Since evaluating all event pairs is computationally prohibitive, we first retrieve candidates for $B$ using \texttt{all-MiniLM-L6-v2} sentence embeddings~\citep{reimers:2019} and select the ten nearest neighbors with cosine similarity of at least 0.3. We additionally propose pairs based on containment of content words. To provide more general concepts that may be absent from the extracted events, we incorporate existing hierarchies from Wikidata~\citep{vrandecic:2014} and the Gene Ontology~\citep{ashburner:2000,thegeneontologyconsortium:2026}, retrieving the three nearest ontology concepts per event. Similarity serves only to retrieve candidates, not to establish subsumption.
For each candidate pair $(A,B)$, a pretrained DeBERTaV3-large NLI model~\citep{he:2023}\footnote{\url{https://hf.co/cross-encoder/nli-deberta-v3-large}}
estimates entailment probabilities in both directions. We infer strict subsumption $A\sqsubset B$ if $P(A \text{\,entails\,} B)\geq0.995$ and $P(B \text{\,entails\,} A)\leq0.10$, and equivalence if both probabilities exceed 0.95. All other pairs remain unconnected. These conservative thresholds prioritize avoiding incorrect generalizations over maximizing coverage.
Finally, we merge mutually entailing descriptions into equivalence groups, limiting their size to avoid excessive merging through potentially incorrect equivalence chains. We then connect the groups through directed subsumption relationships, retaining existing ontology IS-A relationships. We discard contradictory NLI predictions and remove NLI-derived edges involved in cycles to obtain an acyclic graph. Transitive reduction removes redundant edges while preserving multiple parents. Events without accepted relationships remain isolated nodes.

\section{Analysis}
We evaluate the resulting graph on
(1)~extraction quality,
(2)~canonicalization quality,
(3)~hierarchical organization, and
(4)~prevalence of countercausality.
to assess individual pipeline components and illustrate representational capabilities beyond those of the ``flat'' causal graphs from prior work.

\paragraph{Extraction Quality}
We evaluate causality detection (Step 1), candidate extraction, and causality identification (Step 2) separately on the datasets described in Section~\ref{sec:datasets}.
For causality detection, we report binary precision, recall, and F\textsubscript{1}. For candidate extraction, we evaluate span overlap using the character-level Dice coefficient~\citep{dice:1945,liu:2015}. We use optimal one-to-one matching between predicted and annotated spans via the Hungarian algorithm~\citep{kuhn:1955}, which prevents multiple predictions from receiving credit for the same gold span. We aggregate overlap scores and span counts across the test set to compute micro-averaged precision, recall, and F\textsubscript{1}. For causality identification, we report macro-averaged precision, recall, and F\textsubscript{1} over the datasets' classes: \emph{causal} and \emph{noncausal} for CNCv2 and BECauSE~2.0, and \emph{causal}, \emph{countercausal}, and \emph{uncausal} for CCNC. We evaluate identification on gold event pairs to evaluate the identification model itself without any errors propagating from the prior step.

Table~\ref{tab:extraction-results} shows the results. Detection achieves F\textsubscript{1} scores between 84.6\% and 86.9\% across the three datasets. Candidate extraction achieves character-overlap F\textsubscript{1} scores between 69.8\% and 72.8\%. These results indicate that accurately identifying event spans is a challenging task. Note that this is also the case for humans: annotators frequently disagree on the precise boundaries of causal arguments and event mentions~\citep{colruyt:2019,tan:2023}. For our pipeline, exact span boundaries are less important than correctly identifying the participating events. Subsequent canonicalization can use the source text to recover omitted context and construct more complete, self-contained event descriptions. Identification achieves macro-averaged F\textsubscript{1} scores between 89.2\% and 97.4\% on gold event pairs. Effectiveness is lowest on CCNC, which is the only dataset that annotates countercausality. Differences between datasets may also reflect their linguistic characteristics and annotation distributions. Manual inspection of extracted spans revealed that extraction sometimes failed to distinguish epistemic from ontological causality. For example, from ``\textit{Ultra-slow earthquake indicates deep crustal movement},'' the pipeline incorrectly extracts an ontological causal relationship from \emph{``ultra-slow earthquake''} to \emph{``deep crustal movement''}. The title instead expresses epistemic causality: because the earthquake is observed, crustal movement may be inferred.

\begin{table}
\small\centering
\renewcommand{\arraystretch}{.9}%
\belowrulesep=\dimexpr\arraystretch \belowrulesep\relax%
\aboverulesep=\dimexpr\arraystretch \aboverulesep\relax%
\belowbottomsep=\belowrulesep%
\newcommand{\algn}[2]{\hspace{\widthof{0} * #1 + \widthof{\,} * \numexpr(#1-1)/3\relax}\makebox[0pt][r]{#2}}
\caption{Evaluation results (in percent) of our model per dataset and causality extraction subtask.}
\begin{tabular}{@{}p{2cm}*{3}{ccc}@{}}
\toprule
\textbf{Dataset} & \multicolumn{3}{c}{\bfseries Detection} & \multicolumn{3}{c}{\bfseries Candidate Extraction} & \multicolumn{3}{c@{}}{\bfseries Identification} \\
\cmidrule(lr){2-4}\cmidrule(lr){5-7}\cmidrule(l){8-10}
            & P & R & F\textsubscript{1} & P & R & F\textsubscript{1} & P & R & F\textsubscript{1} \\
\midrule
CCNC        & \algn{2}{90}.6 & \algn{2}{83}.6 & \algn{2}{86}.9 & \algn{2}{68}.2 & \algn{2}{74}.1 & \algn{2}{71}.0 & \algn{2}{90}.8 & \algn{2}{88}.1 & \algn{2}{89}.2 \\
CNCv2       & \algn{2}{84}.3 & \algn{2}{87}.0 & \algn{2}{85}.6 & \algn{2}{67}.2 & \algn{2}{79}.4 & \algn{2}{72}.8 & \algn{2}{93}.7 & \algn{2}{93}.6 & \algn{2}{93}.7 \\
BECauSE 2.0 & \algn{2}{78}.6 & \algn{2}{91}.7 & \algn{2}{84}.6 & \algn{2}{65}.4 & \algn{2}{74}.8 & \algn{2}{69}.8 & \algn{2}{96}.7 & \algn{2}{98}.3 & \algn{2}{97}.4 \\
\bottomrule
\end{tabular}
\label{tab:extraction-results}
\end{table}

\paragraph{Canonicalization Quality}
We assess the three canonicalization approaches described in Section~\ref{sec:pipeline}: syntactic expansion, context-aware canonicalization, and LLM-based rewriting. To that end, we compare the event descriptions before and after canonicalization along three dimensions: \emph{self-con\-tain\-ment}, whether an event can be understood independently of its source text; \emph{completeness}, whether the description includes all and only the information relevant to the event; and \emph{faithfulness}, whether it preserves the event's original meaning.
These criteria capture different aspects of event representation. Consider the title ``\emph{Altitude adaptation in Tibetans caused by introgression of Denisovan-like DNA}.'' For the effect event, both ``\emph{altitude adaptation}'' and ``\emph{altitude adaptation in Tibetans}'' are self-contained, since they can be understood without the source title. Both are also faithful, as neither changes the meaning of the described event. However, only the latter is complete: the former omits that the adaptation concerns Tibetans. Thus, an event description can be self-contained and faithful while omitting relevant information from its source.

We conduct a pairwise human evaluation on 50 sampled event spans for which the pre-canon\-i\-cal\-ized and three post-canonicalized event descriptions differ. For each span, we compare all four descriptions in pairs, excluding identical descriptions. This yields 251 comparisons. Annotators are shown the source title, the extracted span, and two descriptions in randomized order. For each criterion, they select the better description or indicate that both are equally good or dissatisfying the criterion. By focusing on spans for which the approaches produce different descriptions, the evaluation concentrates on cases where canonicalization choices make a difference. Consequently, the sample is not representative of all extracted events.

\begin{table}
\small\centering
\renewcommand{\arraystretch}{1}%
\belowrulesep=\dimexpr\arraystretch \belowrulesep\relax%
\aboverulesep=\dimexpr\arraystretch \aboverulesep\relax%
\belowbottomsep=\belowrulesep%
\newcommand{\algn}[2]{\hspace{\widthof{0} * #1 + \widthof{\,} * \numexpr(#1-1)/3\relax}\makebox[0pt][r]{#2}}
\tabcolsep=3.96pt
\caption{(a) Preference rating across the canonicalizers and inter-annotator agreement (Krippendorff's~$\alpha$), and (b) deviation from the extracted span (portion of spans that were changed at all, and mean normalized Levenshtein distance).}
\vspace{-1.25em}
\begin{tabular}{@{}l@{}p{2\tabcolsep}@{}ccc@{}p{2\tabcolsep}@{}cc@{}}
                           && \multicolumn{3}{@{}l@{}}{(a)} && \multicolumn{2}{@{}l@{}}{(b)} \\
\addlinespace[-\aboverulesep]
\cmidrule[\heavyrulewidth]{1-1}
\cmidrule[\heavyrulewidth]{3-5}
\cmidrule[\heavyrulewidth]{7-8}
\textbf{Approach}          && \textbf{Faithfulness} & \textbf{Completeness} & \textbf{Self-containment} && \textbf{Changed (\%)} & \textbf{Norm. edit dist.}\\
                           && ($\alpha=0.785$) & ($\alpha=0.700$) & ($\alpha=0.699$) \\
\cmidrule{1-1}\cmidrule{3-5}\cmidrule{7-8}
Passthrough                && 0.\algn{2}{82} & 0.\algn{2}{73} & 0.\algn{2}{38} && \algn{2}{0}.0 & 0.\algn{2}{00}\\
Syntactic Expansion        && 0.\algn{2}{65} & 0.\algn{2}{60} & 0.\algn{2}{27} && \algn{2}{8}.0 & 0.\algn{2}{04}\\
Context-aware              && 0.\algn{2}{57} & 0.\algn{2}{71} & 0.\algn{2}{65} && \algn{2}{22}.9 & 0.\algn{2}{07}\\
LLM-based rewriting        && 0.\algn{2}{04} & 0.\algn{2}{04} & 0.\algn{2}{61} && \algn{2}{75}.8 & 0.\algn{2}{48}\\
\cmidrule[\heavyrulewidth]{1-1}
\cmidrule[\heavyrulewidth]{3-5}
\cmidrule[\heavyrulewidth]{7-8}
\addlinespace[-\belowrulesep]
\end{tabular}
\label{tab:canonization-results}
\end{table}

Results are shown in Table~\ref{tab:canonization-results}. No approach performs best across all three criteria. This highlights the trade-off between producing self-contained event descriptions and preserving their original meaning. The pre-canonicalized spans achieve the highest faithfulness~(0.82) and completeness~(0.73), but relatively low self-containment~(0.38). Syntactic expansion performs worse across all three criteria, suggesting that expanding phrase boundaries alone is insufficient to recover the relevant context. Context-aware canonicalization achieves the highest self-containment~(0.65) while maintaining similar completeness~(0.71), but at the cost of lower faithfulness~(0.57). It modifies 22.9\% of event descriptions. The rule-based canonicalization approaches do not modify the event descriptions drastically with a mean normalized edit distance of 0.04 and 0.07 respectively.
LLM-based rewriting achieves relatively high self-containment (0.61), but performs poorly on faithfulness and completeness (both 0.04). Manual inspection reveals substantial semantic distortions as LLM-based rewriting introduces too much context information. Sometimes the effect is added to the cause span, e.g., ``\emph{Tiny Flexible Sensors}'' is expanded to ``\emph{A tiny flexible sensor enables artificial retinas}''. Sometimes it hallucinates information, e.g., ``\emph{Dogs are being trained to sniff your urine for cancer}'' is rewritten to ``\emph{Dogs are being trained to sniff out explosives}''. Overall, the results favor context-aware canonicalization over unrestricted LLM rewriting for constructing independently interpretable event descriptions in our pipeline. Nevertheless, preserving their original meaning remains challenging even for rule-based methods.

\paragraph{Hierarchical Representation}
The hierarchy contains \numhiernodes{} nodes connected by \numhieredges{} direct subsumption relationships. The deepest branch has a length of \hierdepthall{}. Overall, \hierlinkedpct{}\% of extracted event descriptions participate in at least one subsumption relationship. The remaining event descriptions are kept as isolated nodes. Table~\ref{tab:hierarchy-results} summarizes the hierarchy's structure and shows the 5 largest nodes (by number of descendants) with a randomly chosen example path from their subhierarchy.
Note that the inferred subsumptions remain subject to known limitations of NLI models like a reliance on spurious cues rather than semantic entailment~\citep{mccoy:2019}. For example, ``\textit{Scientists believe psilocybin might produce rapid and lasting antidepressant effects}'' is said to entail ``\textit{psilocybin might produce rapid and lasting antidepressant effects}'', though just because scientists believe something does not mean that it is true. While the inferred hierarchy is therefore expected to contain noisy subsumption relationships, it nevertheless provides a useful hierarchical signal that is missing from current causal graphs.

\begin{table}[t]
\small\centering
\setlength{\tabcolsep}{4pt}
\renewcommand{\arraystretch}{.9}%
\belowrulesep=\dimexpr\arraystretch \belowrulesep\relax%
\aboverulesep=\dimexpr\arraystretch \aboverulesep\relax%
\belowbottomsep=\belowrulesep%
\caption{Characteristics of the inferred hierarchy and examples from its largest subhierarchies. Paths illustrate inferred subsumption relationships, including erroneous links.}
\label{tab:hierarchy-results}
\renewcommand\example[1]{\multicolumn{3}{@{\hspace{1em}}p{\linewidth-1em}@{}}{%
  \textcolor{gray}{\footnotesize#1}}}
\vspace{-1em}
\begin{tabular}{@{}llr@{}}
\toprule
\multicolumn{2}{@{}l}{\textbf{Hierarchy statistics}}
  & \textbf{Value} \\
\midrule

\multirow{3}{*}{\rotatebox{90}{Nodes}}
& Distinct event descriptions
  & \numhierdescriptions \\
& Event clusters (equivalent descriptions merged)
  & \numhiereventnodes \\
& Ontology concepts without own events
  & \numhierconceptnodes \\

\midrule
\multirow{11}{*}{\rotatebox{90}{Subsumption}}
& Total relationships
  & \numhieredges \\
& \quad Inferred by NLI
  & \numhieredgesnli \\
& \quad Taken from the ontologies
  & \numhieredgesonto \\
& Event $\rightarrow$ event edges
  & \numhieredgesee \\
& Event $\rightarrow$ ontology concept edges
  & \numhieredgesec \\
& Additional (second) parents
  & \numhieredgessecondary \\
& Deepest branch (all parents)
  & \hierdepth{} (\hierdepthall) \\
& Depth of linked events, mean / median
  & \hierdepthmean{} / \hierdepthmedian \\
& Children per inner node, mean / median / max
  & \hierbranchmean{} / \hierbranchmedian{} / \hierbranchmax \\
& Root nodes with children  & \numhierroots \\
& Event descriptions linked & \hierlinkedpct\% \\

\midrule
\multicolumn{2}{@{}l}{\textbf{Largest roots and example paths} {\smaller (nodes extracted from r/science are \underline{underlined})}} & \textbf{\# Desc.} \\
\midrule

\multicolumn{2}{@{}l}{biological process} & 14\,835 \\
\example{reproductive process $\sqsupset$ embryo implantation $\sqsupset$ \uline{deceased donor's uterus} $\sqsupset$ \uline{The First US Uterus Transplant Failed}} \\\noalign{\vspace{\belowrulesep}}

\multicolumn{2}{@{}l}{conflict}  & 1\,913 \\
\example{combat $\sqsupset$ blood sport $\sqsupset$ animal-baiting $\sqsupset$ insect fighting $\sqsupset$ \uline{to kill mosquitoes} $\sqsupset$ \uline{to trap and kill mosquitoes}} \\\noalign{\vspace{\belowrulesep}}

\multicolumn{2}{@{}l}{disaster}  & 1\,744 \\
\example{cataclysm $\sqsupset$ extinction event $\sqsupset$ \uline{The asteroid strike that killed off the dinosaurs occurs.} $\sqsupset$ \uline{A massive asteroid strike that wiped out the dinosaurs millions of years ago created room for frogs to colonize the Earth, occurs.}} \\\noalign{\vspace{\belowrulesep}}

\multicolumn{2}{@{}l}{crisis} & 824 \\
\example{emergency $\sqsupset$ distress $\sqsupset$ mental distress $\sqsupset$ \uline{patients with depression} $\sqsupset$ \uline{clinical mood recovery} $\sqsupset$ \uline{Patients with moderate to severe depression reported significant improvements in mood}} \\\noalign{\vspace{\belowrulesep}}

\multicolumn{2}{@{}l}{accident} & 750 \\
\example{fall $\sqsupset$ falling from the sky $\sqsupset$ \uline{a meteor fall} $\sqsupset$ \uline{Two meteorites fell down in India}} \\\noalign{\vspace{\belowrulesep}}
\bottomrule
\end{tabular}

\end{table}

\paragraph{Causal Counterclaims}
Of the \numafterextraction{} extracted relationships, \numcountercausal{} (\countercausalpct{}\%) are countercausal. Their prevalence varies across scientific domains: \emph{\countercausalfieldhigh{}} has the highest relative share of countercausal claims (\countercausalfieldhighpct{}\%) and \emph{\countercausalfieldlow{}} the lowest (\countercausalfieldlowpct{}\%). After canonicalization and equivalence grouping, \numconflictpairs{} directed event pairs are both in the causal and countercausal relation. Additionally, we identify \numconflicttriples{} triples $(A,B,C)$ with $B\sqsubset A$, where $(A,C)$ and $(B,C)$ occur with opposite causal polarities. Unlike directly opposing claims about the same event pair, these cases may reflect more specific exceptions to general causal assertions rather than outright contradictions. For example, the claims ``\textit{Exercise Can Prevent Depression}'' (\href{https://www.reddit.com/r/science/comments/aj4yx7}{r/science/aj4yx7}) and ``\textit{Exercise `no help for depression'{}}'' (\href{https://www.reddit.com/r/science/comments/unl72}{r/science/unl72}) illustrate potentially opposing assertions across sources. Such cases illustrate how the combination of counterclaims and hierarchical representation makes potentially conflicting causal knowledge accessible across different levels of specificity.
Counterclaims are also important for representing disputed and revised causal knowledge. For example, CauseNet~\citep{heindorf:2020} and Cause Effect Graph~\citep{li:2020} contain the assertion that vaccines cause autism, although extensive scientific evidence does not support this relationship~\citep{worldhealthorganization:2026}. Our graph preserves a corresponding counterclaim (\href{https://www.reddit.com/r/science/comments/axd4l4}{r/science/axd4l4}). Similarly, both resources contain the assertion that sugar consumption causes hyperactivity, a historically widespread belief but controlled studies find no overall behavioral effect~\citep{wolraich:1995}. Our graph also contains a corresponding counterclaim (\href{https://www.reddit.com/r/science/comments/6m8cr}{r/science/6m8cr}). These examples illustrate how positive-only causal graphs can retain disputed or outdated assertions while omitting evidence that challenges them. By preserving counterclaims and their provenance, our representation makes such opposing assertions accessible without treating extracted relationships as established causal facts.

\section{Discussion}
Our resource builds on prior causal claim graph extraction and improves upon it by jointly representing causal claims and counterclaims, preserving their provenance, and organizing canonicalized event descriptions hierarchically. Context-aware canonicalization produces more self-contained event descriptions. Hierarchical representation connects events at different levels of specificity without conflating their causal relationships. Together, these features allow potentially conflicting claims to be retrieved and compared across sources, without imposing a single causal interpretation.
The graph can provide candidate background knowledge for causal discovery~\citep{bang:2025} and support approaches that consider uncertain or inconsistent causal information~\citep{padh:2025,baldo:2026}. We stress that the extracted claims do not establish causal relationships. Using them as constraints on causal models needs more evidence and assumptions, including mapping textual events to model variables. Similarly, the hierarchy may support reasoning over general claims and specific exceptions~\citep{pearl:1988,bochman:2023}, but does not resolve conflicting claims.

Like other automatically extracted causal resources, our graph contains false positives and other errors. Nevertheless, previous work has demonstrated applicability of such resources, including to causal question answering over CauseNet~\citep{blubaum:2024}. Our analysis identified errors in causality extraction, semantic changes during canonicalization, and incorrect subsumption relationships. These errors can propagate through the pipeline, potentially affecting the interpretation and comparison of claims. The resulting graph should therefore be understood as a source of candidate causal knowledge rather than a collection of verified causal facts.
Further limitations arise from the source data. Reddit submission titles summarize scientific findings, but may omit relevant conditions, populations, and methodological qualifications. As such, the graph represents claims reported in these titles, which may not reflect the precise conclusions of the underlying studies. However, the provenance information allows claims to be traced to their sources.

\section{Conclusion}
Much potentially useful causal knowledge is expressed in natural language, which makes it difficult to access for systematic causal reasoning. We introduced a large-scale graph of \numaftercanonicalization{} causal and countercausal claims extracted from \numtitles{} Reddit r/science submission titles. Unlike existing large-scale causal graphs, our resource jointly preserves counterclaims, textual provenance, and a hierarchy of canonicalized event descriptions. Our extraction models achieve F\textsubscript{1} scores of 84.6--86.9\% for causality detection and 89.2--97.4\% for causality identification on held-out datasets. Canonicalization improves the self-containment of event descriptions, although preserving their meaning remains challenging. The inferred hierarchy connects 61.5\% of event descriptions, while the graph captures opposing claims both between the same event pairs and across different levels of specificity. Together, these features make reported causal relationships accessible for further analysis and reasoning without imposing a single causal model.
Future work should address semantically challenging cases, including the distinction between epistemic and ontological causality and statements that question the evidence for a causal relationship. Further improvements include representing uncertainty and hedging in causal claims, such as \textit{``may cause''} or \textit{``could cause,''} and extending the pipeline to larger and more diverse textual sources. Finally, we see promise in investigating how the resulting graph supports causal question answering, hypothesis formation, and qualitative neuro-symbolic reasoning over incomplete and conflicting causal knowledge.

\acks{Liesbeth Allein is supported by a junior postdoctoral fellowship from Research Foundation - Flanders (FWO) under grant 12AGW26N.}

\bibliography{arxiv27-causality-rscience-lit}

\appendix
\section{Dataset and Extraction Evaluation}
\paragraph{Dataset Statistics}
We train and evaluate the causality extraction models on CCNC~\citep{hagen:2025a}, CNCv2~\citep{tan:2023}, and BECauSE 2.0~\citep{dunietz:2017}. Table~\ref{tab:dataset-statistics} summarizes their sizes and class distributions. For CCNC and CNCv2, we use the official validation sets as held-out test sets and split the original training data into new training and validation partitions. For BECauSE 2.0, we follow the train/test split used by UniCausal~\citep{tan:2023a} and reserve 10\% of the training data for validation.

\begin{table}
\small\centering
\def\arraystretch{0.9}
\tabcolsep=2.5pt
\newcommand{\algn}[2]{\hspace{\widthof{0} * #1}\makebox[0pt][r]{#2}}%
\caption{Dataset split sizes of the datasets used for training and evaluation. For detection, y, refers to the number of samples that contain causal information and n for those that do not. For candidate extraction, the number of spans is given, and for identification the distribution of causal (c), countercausal (cc), and uncausal (uc) samples.}
\begin{tabular}{@{}l*{3}{c@{/}c}*{3}{c}*{3}{c@{/}c@{/}c}@{}}
\toprule
\textbf{Dataset} & \multicolumn{6}{c}{\textbf{Detection}} & \multicolumn{3}{c}{\textbf{Candidate Extraction}} & \multicolumn{9}{c}{\textbf{Identification}} \\
\cmidrule(lr){2-7}\cmidrule(lr){8-10}\cmidrule(l){11-19}
& \multicolumn{2}{c}{Train} & \multicolumn{2}{c}{Val.} & \multicolumn{2}{c}{Test}
& \multicolumn{1}{c}{Train} & \multicolumn{1}{c}{Val.} & \multicolumn{1}{c}{Test}
& \multicolumn{3}{c}{Train} & \multicolumn{3}{c}{Val.} & \multicolumn{3}{c@{}}{Test}\\
\cmidrule(lr){2-3}\cmidrule(lr){4-5}\cmidrule(lr){6-7}
\cmidrule(lr){11-13}\cmidrule(lr){14-16}\cmidrule(l){17-19}
                & y & n & y & n & y & n & \#spans & \#spans & \#spans & c  &  cc &  uc &  c  &  cc &  uc &  c  &  cc & uc  \\
\midrule
CCNC             & \algn{4}{1588} & \algn{4}{1179} & \algn{3}{185} & \algn{3}{123} & \algn{3}{207} & \algn{3}{133} & \algn{4}{3771} & \algn{3}{436} & \algn{3}{473} & \algn{4}{1135} & \algn{3}{765} & \algn{4}{4290} & \algn{3}{130} & \algn{2}{90} & \algn{3}{472} & \algn{3}{117} & \algn{3}{117} & \algn{3}{512} \\
CNCv2            & \algn{4}{1455} & \algn{4}{1312} & \algn{3}{169} & \algn{3}{139} & \algn{3}{185} & \algn{3}{155} & \algn{4}{3842} & \algn{3}{442} & \algn{3}{484} & \algn{4}{2026} & \algn{3}{0} & \algn{4}{5880} & \algn{3}{231} & \algn{2}{0} & \algn{3}{697} & \algn{3}{249} & \algn{3}{0} & \algn{3}{769} \\
BECauSE      & \algn{4}{273} & \algn{4}{134} & \algn{3}{33} & \algn{3}{13} & \algn{3}{12} & \algn{3}{5} & \algn{4}{737} & \algn{3}{83} & \algn{3}{28} & \algn{4}{377} & \algn{3}{0} & \algn{4}{1259} & \algn{3}{43} & \algn{2}{0} & \algn{3}{109} & \algn{3}{14} & \algn{3}{0} & \algn{3}{30} \\
\bottomrule
\end{tabular}
\label{tab:dataset-statistics}
\end{table}

\begin{figure}
\centering
\includegraphics{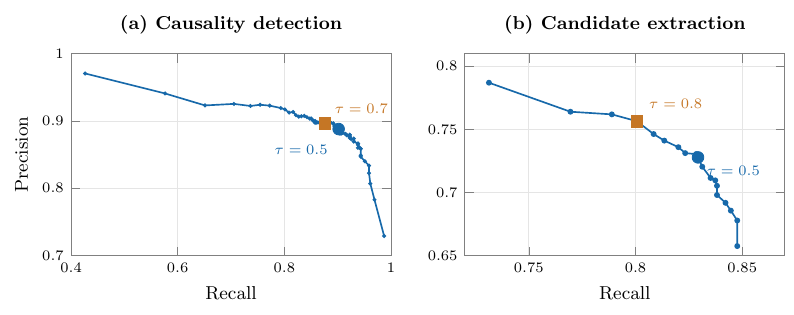}
\caption{The precision-recall curves of the trained model on the pooled validation splits of CCNC, CNCv2, and BECauSE 2.0 on (a)~the causality detection and (b)~the candidate extraction subtask respectively. Each dot marks a separate decision threshold. The knee in both cases is around 0.5.}
\label{fig:detection-pr}\label{fig:candidate-extraction-pr}
\end{figure}

\paragraph{Threshold Selection}
\label{sec:detection-threshold}\label{sec:candidate-extraction-threshold}
We select the detection and candidate extraction thresholds based on pre\-ci\-sion--recall analysis on the validation data. Figure~\ref{fig:detection-pr} (a) shows the trade-off for causality detection. We retain a threshold of 0.5 to prioritize recall, since titles discarded during detection cannot be recovered by subsequent pipeline steps. On the CCNC validation set, increasing the threshold to 0.7 reduces recall from 0.892 to 0.881.
For candidate extraction, we select a span confidence threshold of 0.5 based on character-overlap precision and recall (Figure~\ref{fig:candidate-extraction-pr} (b)). Increasing the threshold to 0.8 improves precision from 0.728 to 0.756 but reduces recall from 0.829 to 0.801, while F1 changes only marginally (0.775 to 0.778). We therefore retain the lower threshold to recover more candidate event spans for subsequent identification and canonization.

\section{Event Canonization}
\paragraph{LLM-based Canonization}
\label{sec:llm-canonization}
In addition to the syntactic and context-aware canonizers, we evaluate event rewriting with Qwen2.5-3B-Instruct. Given a source title and a marked event span, the model is instructed to produce a self-contained description while preserving the event's original meaning and excluding information belonging to the paired event. The full prompt is shown in Listing~\ref{lst:prompt}. We use constrained decoding with a JSON output schema and seven few-shot examples.
Generated descriptions are checked for source faithfulness and leakage from the paired event. In particular, the automated checks require at least 60\% of the source content lemmas to be retained and reject descriptions that incorporate content from the paired event. In the graph released with this paper, we retain the deterministic context-aware canonizer's descriptions; LLM-generated descriptions are used for the comparative evaluation.

\begin{lstlisting}[style=prompt, float=t, caption={The few-shot prompt we used for canonicalization. To canonicalize the text, \texttt{<CONTEXT>} and \texttt{<PHRASE>} are replaced with the actual context sentence and phrase.}, label={lst:prompt}]
{"role": "system", "content": "You rewrite a short phrase describing an event into ONE self-contained sentence describing ONLY that event. The Context describes a cause-and-effect relationship, but the Phrase is only ONE side of it (either the cause or the effect, not both). Your sentence must NOT state what the phrase's event causes, what caused it, or the outcome/result described elsewhere in the Context - describe the event or concept in the Phrase in isolation, as if the rest of the causal claim did not exist. If the phrase alone is too short or generic to form a sentence (e.g. a single name), you may pull a detail from Context about what THIS thing is or how it was used/administered/observed - never the outcome or consequence attributed to it. Use only facts stated in the phrase or context; do not add information that isn't there. Output JSON with a single 'sentence' field."}
{"role": "user", "content": "Context: Chronic sleep deprivation leads to cognitive decline in the elderly with dementia.\nPhrase: cognitive decline in the elderly with dementia"}
{"role": "assistant", "content": "{\"sentence\": \"Cognitive decline occurs in the elderly with dementia.\"}"}
{"role": "user", "content": "Context: Excessive metal mining depletes natural mineral reserves.\nPhrase: the mining of metals"}
{"role": "assistant", "content": "{\"sentence\": \"Metals are mined.\"}"}
{"role": "user", "content": "Context: Genetically Engineered Immune Cells Found to Rapidly Clear Leukemia Tumors.\nPhrase: Genetically Engineered Immune Cells"}
{"role": "assistant", "content": "{\"sentence\": \"Immune cells are genetically engineered.\"}"}
{"role": "user", "content": "Context: Cancer drug aids the regeneration of spinal cord injuries.\nPhrase: aids the regeneration of spinal cord injuries"}
{"role": "assistant", "content": "{\"sentence\": \"Spinal cord injuries are regenerated.\"}"}
{"role": "user", "content": "Context: Ayahuasca administration was found to increase serotonin and dopamine turnover in the rat brain compared to control.\nPhrase: Ayahuasca"}
{"role": "assistant", "content": "{\"sentence\": \"Ayahuasca was administered to rats.\"}"}
{"role": "user", "content": "Context: A new grant boosts research funding for renewable energy projects.\nPhrase: A new grant"}
{"role": "assistant", "content": "{\"sentence\": \"A new grant was awarded.\"}"}
{"role": "user", "content": "Context: A new grant boosts research funding for renewable energy projects.\nPhrase: boosts research funding for renewable energy projects"}
{"role": "assistant", "content": "{\"sentence\": \"Research funding for renewable energy projects increases.\"}"}
{"role": "user", "content": "Context: <CONTEXT>\nPhrase: <PHRASE>"}
\end{lstlisting}

\end{document}